# Quaternion-Hadamard Network: A Novel Defense Against Adversarial Attacks

Vladimir Frants, Sos Agaian, *Life Fellow, IEEE*

***Abstract*—Adverse-weather image restoration (e.g., rain, snow, haze) models remain highly vulnerable to gradient-based white-box adversarial attacks, wherein minimal loss-aligned perturbations cause substantial degradation in the restored output. This paper presents QHNet, a computationally efficient purification-based defense that precedes the restoration network and targets perturbation suppression in the transform and quaternion domains. QHNet incorporates a Quaternion Hadamard Polynomial Denoising Block (QHPDB) and a Quaternion Denoising Residual Block (QDRB) within an encoder–decoder framework to remove high-frequency adversarial noise while preserving fine structural details. Robustness is evaluated using PSNR and SSIM across rain, snow, and haze removal tasks, and further validated under adaptive, defense-aware white-box attacks employing Projected Gradient Descent (PGD), Backward Pass Differentiable Approximation (BPDA), and Expectation Over Transformation (EOT). Experimental results demonstrate that QHNet delivers superior restoration fidelity and significantly improved robustness compared to state-of-the-art purification baselines, confirming its effectiveness for low-level vision pipelines.**

***Index Terms*—Hadamard Transform, Quaternion Neural Network, Computer Vision, Image Processing**

## I. INTRODUCTION

The rise of autonomous driving and advanced surveillance systems underscores the importance of robustness and efficiency in adverse weather conditions. State-of-the-art rain, snow, and haze removal techniques can significantly improve image quality, enhancing the visibility of details [1], [2]. Deep learning models for image processing, including adverse-weather removal, achieve strong performance but are highly vulnerable to adversarial attacks [3]. These attacks introduce small, often imperceptible perturbations that can mislead the model and cause severe failures [4], [5]. As a result, adversarial noise poses a serious threat to both the visual quality and the practical reliability of weather-removal systems. Adversarial perturbations can severely disrupt weather-removal models, as illustrated in Fig. 1: (a) the network may fail to remove rain, haze, or snow; (b) the restored image may contain strong artifacts or unnatural patterns; and (c) the scene may undergo major distortions or semantic changes. Such failures significantly compromise downstream tasks like autonomous driving, surveillance, and remote sensing. Although various defenses have been proposed [6]–[11], many are computationally heavy, vulnerable to adaptive attacks, and primarily evaluated on classification rather than restoration quality. Meanwhile, recent compact architectures, MobileNets, attention-based models, transform-domain networks, capsule networks, and quaternion neural networks (QNNs) [12]–[17]—offer efficiency but have largely untested robustness. QNNs are particularly appealing for color image restoration because they jointly process RGB channels via the Hamilton product, exploit inter-channel correlations, reduce parameters by up to 4×, and improve robustness to perturbations [18], [19].

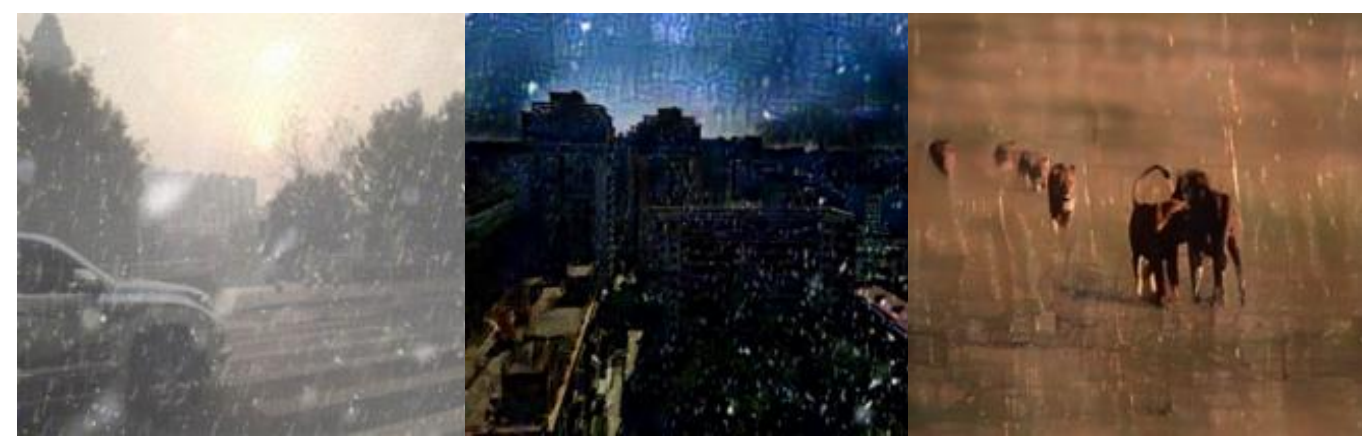

a) b) c)

**Fig. 1.** Effects of adversarial attacks on weather removal methods. (a) inability to remove the weather condition; (b) severe artifacts; (c) severe image alteration.

TABLE I: DEFENSES AGAINST ADVERSARIAL ATTACKS

| **Method** | **Training** | **Artifacts** | **Effectiveness** |
|---|---|---|---|
| Distillation [4], [7], [20] | Yes | Low | High |
| JPEG compression [8] | No | Low | Moderate |
| Input transformations [9] | No | Moderate | High |
| Pixel deflection [10] | No | Moderate | High |
| Inpainting [11] | No | High | High |
| Super-resolution [21] | No | High | Low |
| Purification (GAN, PixelCNN, Diffusion) [6], [22]–[24] | Yes | Moderate | High |

To address these limitations, we introduce the Quaternion–Hadamard Network (QHNet)**,** a lightweight purification-based defense that neutralizes adversarial perturbations before they enter the restoration model. Our work specifically targets gradient-based white-box attacks**,** which constitute the strongest and most damaging threat model. In this setting, the adversary possesses full knowledge of both the restoration network and the defense, and seeks the minimal perturbation aligned with the model's loss gradient to maximize degradation of the output. These gradient-aligned perturbations, although visually subtle, can severely distort restored images and compromise downstream decision-making. The key contributions of this work are as follows:

1. Polynomial Thresholding Layer: We introduce a polynomial thresholding layer that operates in the Walsh–Hadamard Transform (WHT) domain to enhance perturbation suppression and reduce susceptibility to gradient-based adversarial attacks. The layer is robust to Gaussian noise, BPDA, and EOT-based adaptive attacks, and enforces structured shrinkage of high-frequency components where adversarial perturbations tend to concentrate.

2. Quaternion–Hadamard Neural Network (QHNet): We propose QHNet, an efficient purification architecture that achieves strong adversarial noise reduction at a fraction of the computational cost of diffusion-based defenses. QHNet is composed of three primary modules:

- Quaternion Hadamard Polynomial Denoising Block (QHPDB): Enhances transform-domain denoising using quaternion algebra combined with polynomial thresholding.
- Quaternion Denoising Residual Block (QDRB): Refines feature representations while preserving structural and perceptual integrity.
- Quaternion Feature Aggregation and Refinement Block (QFARB): Aggregates multiscale quaternion features to improve robustness against complex, spatially varying disturbances such as haze, snow, and rain streaks.

3. Comprehensive Defense-Aware Evaluation: We conduct extensive defense-aware experiments across multiple CNN and transformer-based restoration architectures and a broad range of artifact-removal tasks, including Gaussian noise removal, dehazing, deraining, and desnowing. Under fully adaptive white-box conditions, QHNet attains robustness comparable to diffusion-based purification methods while using significantly less computational power, enabling near real-time deployment.

The remainder of this paper is organized as follows: Section II reviews related work; Section III presents the methodology and internal components of QHNet; Section IV describes the dataset construction process; Section V reports comparative results and analyses; and Section VI concludes the paper with key findings and future directions.

## II. Related work

This work focuses on gradient-based white-box attacks, the strongest threat model in which the adversary has full access to the model's architecture, parameters, and training process, enabling precise loss-aligned perturbations. In low-level vision tasks such as dehazing, deraining, and desnowing, the output is a full image rather than a class label, and real paired ground truth is often unavailable. As a result, the dominant practice in the literature is to use first-order, $L_p$-bounded attacks (FGSM, I-FGSM, PGD) optimized with restoration losses or spatial masks rather than classification margins. Prior studies confirm this trend: pseudo-target dehazing attacks [4], region-restricted PGD for deraining [25], degradation-optimized attacks for super-resolution [26], denoising-PGD variants [27], transformer-based restoration vulnerabilities [28], and similar findings in underwater enhancement [29], [30]. Given these constraints, we target robustness within this restoration-specific threat model rather than general-purpose classification defenses.

Recent adoption of Vision Transformers (ViTs) introduces additional vulnerabilities: Aldahdooh et al. [31] show differing $L_p$-norm robustness between ViTs and CNNs, and Mahmood et al. [32] observe low transferability between the two, underscoring the need for defense methods that generalize across architectures.

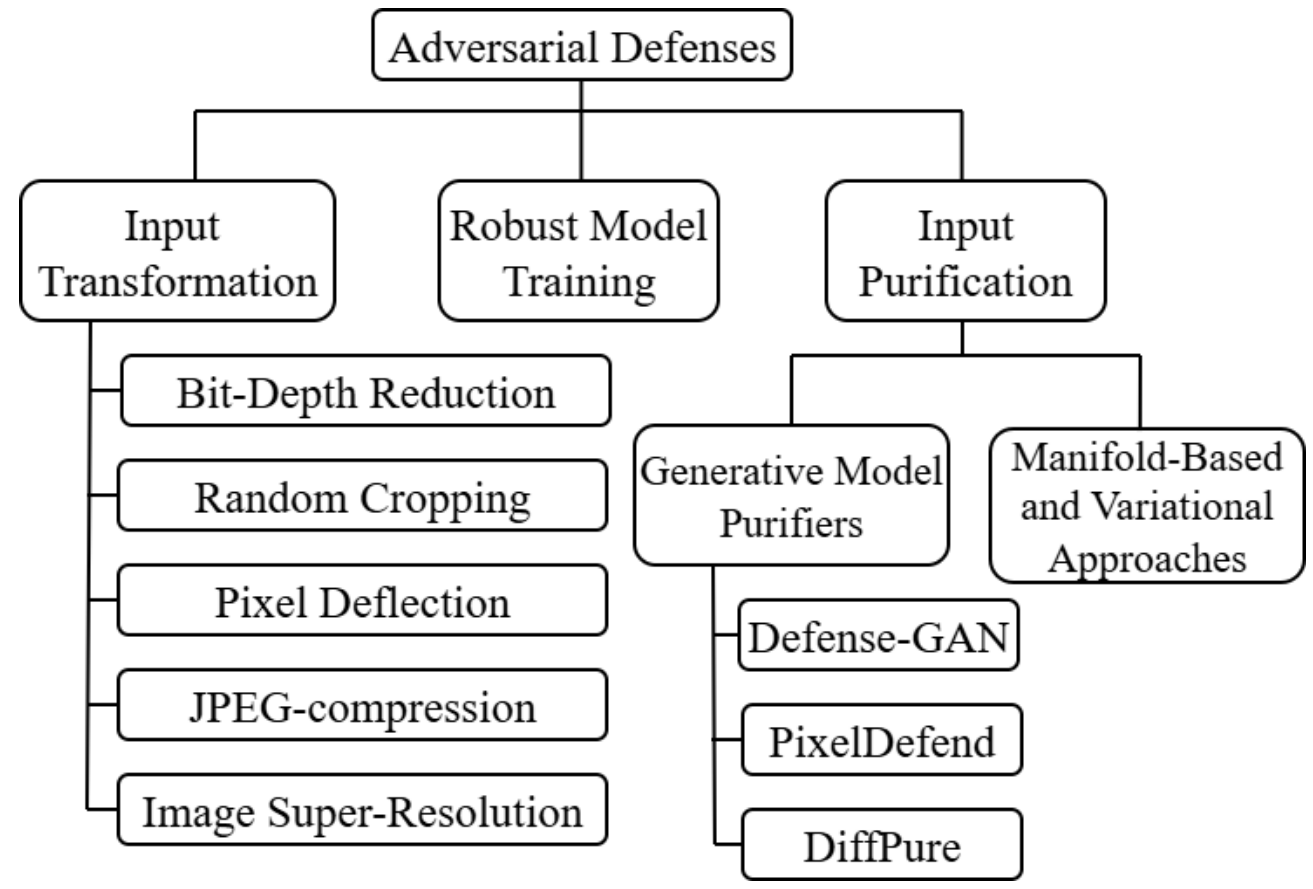

Fig. 2. Taxonomy of adversarial defenses

Existing defenses fall into three categories: (1) robust training, including adversarial training, which improves robustness but is computationally intensive and often harms clean accuracy; (2) input transformations, such as JPEG compression or bit-depth reduction [8]–[11], which are simple but introduce artifacts and are easily bypassed via BPDA [33]; and (3) input purification, ranging from GAN/ VAE-based purifiers (DefenseGAN [22], PixelDefend [6]) to modern score-based and diffusion models (DiffPure [23], energy-guided approaches [34], and ADBM [24]). While effective, these often suffer from robustness–fidelity trade-offs and require careful adaptive evaluation. Manifold- and VAE-based approaches [35], along with randomized smoothing [38], offer additional perspectives but still face practical limitations.

Across all categories, a key weakness is gradient masking, where defenses appear robust only because they obstruct gradient computation. Athalye et al. [33] demonstrated that such defenses fail under BPDA and EOT, making them unsuitable for genuine white-box robustness. These limitations motivate the need for lightweight, defense-aware purification methods tailored to low-level vision models.

## III. Proposed method

In the following subsections, we first provide an overview of the proposed QHNet. Then, we introduce the polynomial thresholding (PT) algorithm, Quaternion Hadamard Polynomial Denoising Block (QHPDB), and Quaternion Denoising Residual Block (QDRB). Next, we describe the Quaternion Feature Aggregation and Refinement Block (QFARB). Finally, we discuss the training strategy and model optimization.

### *A. Image Data Representation and Processing*

Quaternion numbers extend the concept of complex numbers to 4 dimensions and can be written as $q = a + bi + cj + dk$, where $a$, $b$, $c$, and $d$ are real numbers, and $i, j$, and $k$ follow these multiplication rules: $i^2 = j^2 = k^2 = ijk = -1$, $ij = k, ji = -k, jk = i, kj = -i, ki = j, ik = -j$ [36]. The input image $\mathrm{I_{in}} \in \mathbb{R}^{M\times N\times 3}$, with color channels (R, G, and B) and spatial dimensions $M \times N$ is encoded using a quaternion-valued matrix:

$$\mathrm{Q} = 0 + \mathrm{R}i + \mathrm{G}j + \mathrm{B}k \quad (1)$$

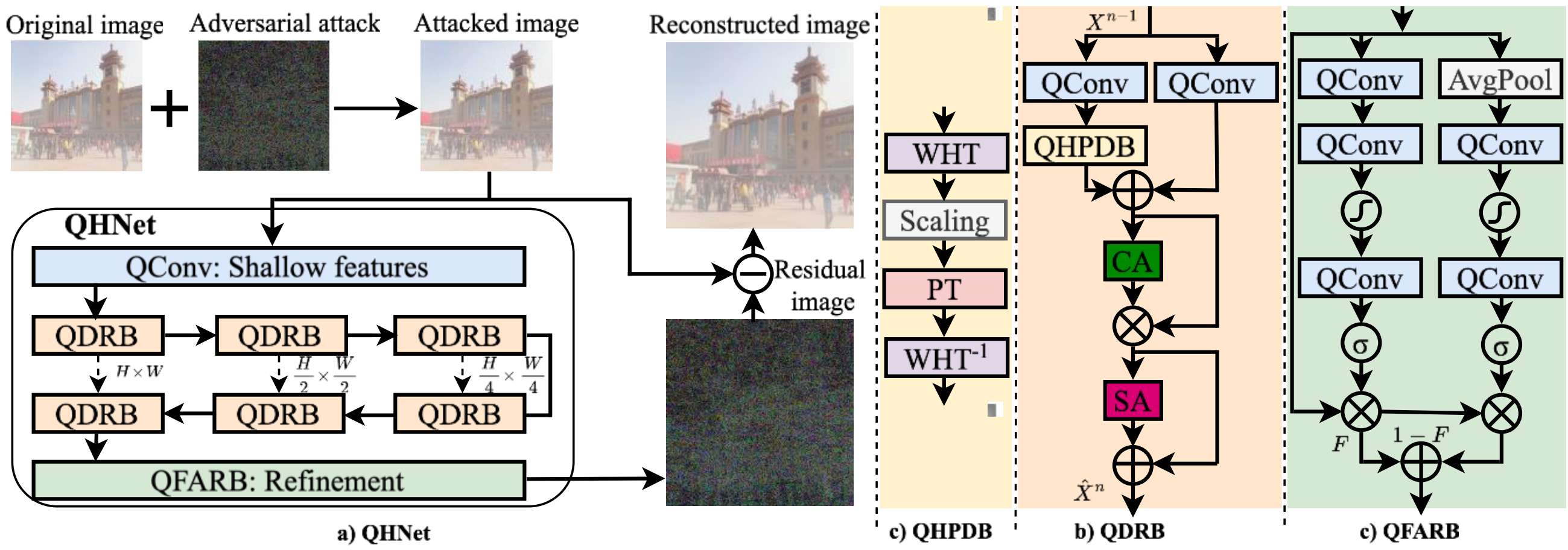

**Fig. 3.** QHNet mitigates adversarial attacks by first transforming the attacked input image into a quaternion representation. It then processes the image through an encoder-decoder architecture built with Quaternion Denoising Residual Block (QDRB), incorporating spatial and channel attention mechanisms. Polynomial thresholding is applied to denoise in the frequency domain. Finally, the Quaternion Feature Aggregation and Refinement Block (QFARB) produces a perturbation-free image that is safe for further processing by the target model.

where $\mathrm{R}, \mathrm{G}, \mathrm{B} \in \mathbb{R}^{M\times N}$ are color channels of the image normalized in the range $[0, 1]$.

Properties of QNNs are defined not by the representation itself, but by how quaternion values are processed. The Hamilton product is used for operations on quaternions. The product of two quaternions $\mathrm{p} = p_r + p_i i + p_j j + p_k k$ and $\mathrm{q} = q_r + q_i i + q_j j + q_k k$ is given by:

$$\begin{aligned} \mathrm{p} \otimes \mathrm{q} = &\left(p_r q_r - p_i q_i - p_j q_j - p_k q_k\right) \\ &+\left(p_r q_i + p_i q_r + p_j q_k - p_k q_j\right) i \\ &+\left(p_r q_j - p_i q_k + p_j q_r + p_k q_i\right) j \\ &+\left(p_r q_k + p_i q_j - p_j q_i + p_k q_r\right) k \end{aligned} \quad (2)$$

The quaternion convolution $\mathrm{QConv}(\mathrm{Q}, \mathrm{K})$ combines the Hamilton product applied pointwise with the usual sliding window operation:

$$(\mathrm{Q} * \mathrm{K})_{(\mathrm{m,n})} = \sum_{\mathrm{u}} \sum_{\mathrm{v}} \left(\mathrm{Q}_{(\mathrm{m+u,n+v})} \otimes \mathrm{K}_{(\mathrm{u,v})}\right) \quad (3)$$

where $\mathrm{Q} = Q_r + Q_i i + Q_j j + Q_k k$ and $\mathrm{K} = K_r + K_i i + K_j j + K_k k$ are quaternion-valued matrices representing the input image and the filter weights, respectively. Here, $m$ and $n$ are the spatial coordinates of the output feature map, while $u$ and $v$ are the spatial coordinates of the filter kernel K.

We use a split-activation function that operates independently on the components of the quaternion-valued feature map. Given a quaternion-valued feature map $\mathrm{Q} = Q_r + Q_i i + Q_j j + Q_k k$, the split-activation function $\varphi$ operates as follows:

$$\mathrm{Q} = \varphi(Q_r) + \varphi(Q_i) i + \varphi(Q_j) j + \varphi(Q_k) k \quad (4)$$

where $\varphi(\cdot)$ is a real-valued activation function.

### *B. QHNet architecture*

The proposed network architecture addresses adversarial attacks using a UNet-like encoding-decoding framework with skip connections (Fig. 3). It starts with a quaternion convolutional layer with a 3x3 kernel to produce shallow features. These features are then processed by groups of K-stacked Quaternion Denoising Residual Blocks (QDRBs) to generate feature maps at full, half, and quarter resolutions. Each QDRB combines a quaternion convolutional layer and a QHPDB for feature extraction and transformation across spatial and frequency domains. This dual-domain processing helps distinguish the original signal from adversarial noise, enabling effective suppression through the Polynomial Thresholding (PT) layer. After decoding, the feature maps are refined by QFARB. The network reconstructs a residual image containing the estimated additive attack noise, which is then subtracted from the original image to produce the final output with suppressed adversarial attack effects.

**Polynomial Thresholding layer (PT):** The polynomial thresholding layer is crucial as an activation function in the frequency domain. Typically, thresholding operators are used for denoising in the wavelet domain through the following steps: (1) orthogonal transform, (2) thresholding, and (3) inverse orthogonal transform. We adopt polynomial thresholding in the WHT domain, using surrogate gradients to achieve smooth gradients during the training phase for effective learning [37], [38]. The layer remains non-differentiable during inference, making the network resistant to gradient-based attacks. Polynomial thresholding generalizes commonly used soft and hard thresholds, providing more flexibility.

The polynomial thresholding operator $T_{\delta,a}(x)$ is defined as follows:

$$T_{\delta,a}(x) = \begin{cases} a_{Z-1} x - a_Z sgn(x)\delta & \text{if } |x| > \delta \\ \sum_{k=0}^{Z-2} a_k x^{2k+1} & \text{if } |x| < \delta \end{cases} \quad (5)$$

Here, $\delta$ is the threshold, $a$ is the vector of polynomial coefficients, $Z$ is the number of terms in the polynomial, and $\mathrm{sgn}(x)$ is the sign function. The general form of the thresholding operator can be expressed in the matrix form:

$$\mathrm{T}_{\delta,\mathrm{a}}(\mathrm{x}) = \mathrm{f}(\mathrm{x}) \cdot \mathrm{a} \quad (6)$$

where $f(x) = [f_0(x), f_1(x), \ldots, f_Z(x)]$ is a vector of functions applied to the input $x$, defined as:

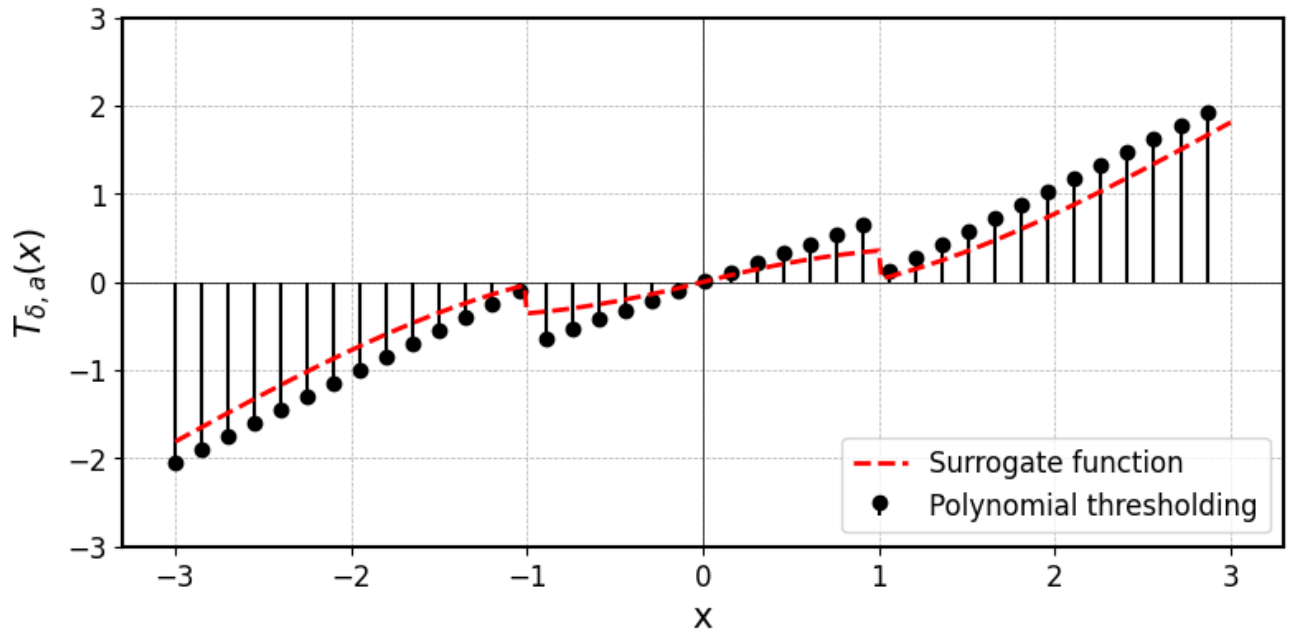

**Fig. 4.** Polynomial thresholding in QHNet. The black curve shows the true operator $T_{\delta,a}(x)$: small Hadamard-domain coefficients ($|x| < \delta$) are reduced by a learnable polynomial, while larger ones undergo linear shrinkage. The red dashed curve is the smooth surrogate used only for backpropagation; inference restores the non-differentiable threshold. This suppresses attack-like high-frequency noise while disrupting gradient-based white-box optimization, without retraining the restoration model.

$$f(x) = \begin{cases} [0, 0, \dots, 0, x - \delta sgn(x)] & \text{if } |x| > \delta \\ [x, x^3, \dots, x^{2Z-3}, 0, 0] & \text{if } |x| < \delta \end{cases} \quad (7)$$

An optimum solution for $a$ can be found by solving the following optimization problem as follows:

$$a_{opt} = \arg\min_{a} \|d - W^T f(Y) a\| \quad (8)$$

where d is the desired attack-free image, $a_{opt}$ is the optimal set of parameters $a$, $W$ is the transform matrix, $Y = W \cdot y$ is the transformed version of the measured image. For an energy-preserving transform such as Walsh-Hadamard, this can be simplified to:

$$a_{opt} = \arg\min_{a} \|D - f(Y) a\| \quad (9)$$

where D is the transformed version of the desired signal d.

When considering many observations, we can alternatively find the minimum MSE (MMSE) error across all the observations:

$$a_{opt} = E\left(f^T(Y) f(Y)\right)^{-1} E\left(f^T(Y) D\right) \quad (10)$$

where $E(\cdot)$ represents the expected value estimation on the whole dataset. For grayscale images attacked with FGSM, $\delta = 1.0$, $Z = 5$ we found $a = [0.707, 0.014, 0.008, 0.999, 0.940]$ (Fig. 4). During the training phase, we replace the hard threshold condition with a sigmoid function, introducing the following surrogate function:

$$T_{\delta,a}(x) = \sigma(|x| - \delta) a_{Z-1} x - \sigma(|x| - \delta) a_Z sgn(x) \delta + \left(1 - \sigma(|x| - \delta)\right) \sum_{k=0}^{Z-2} a_k x^{2k+1} \quad (11)$$

where, σ denotes the sigmoid function, which replaces the traditional hard thresholding condition.

The polynomial thresholding layer is presented in Algorithm 1 and operates by first reshaping the input tensor $X$ of size $B \times C \times M \times N$ into size $B \times C \times M \cdot N$. The tensor of trainable thresholds $\delta$ is then expanded to match the dimensions of $\hat{X}$. Next, the absolute value $|\hat{X}|$ and the sign $sgn(\hat{X})$ of $\hat{X}$ are computed. The condition tensor $O$ is calculated, where each

**Algorithm 1** Polynomial Thresholding Layer

**Require:** Tensor $X \in \mathbb{R}^{B\times C\times M\times N}$, coefficients $a \in \mathbb{R}^{Z}$, threshold $\delta \in \mathbb{R}^{C\times 1}$

Reshape $X$ to $\mathbb{R}^{B\times C\times (M\times N)}$
Expand $\delta$ to $\delta' \in \mathbb{R}^{B\times C\times (M\times N)}$
Compute $|\hat{X}|$ and $sgn(\hat{X})$
$O \leftarrow |\hat{X}| > \delta'$
Initialize $f_X \in \mathbb{R}^{B\times C\times M\times N\times Z}$ to zeros
**for** $i \in \{1, \dots, M \times N\}$ **do**
  **if** $O[i]$ is True **then**
    $f_X[i] \leftarrow [0, 0, \dots, 0,\ \hat{X}[i] - \delta' \cdot sgn(\hat{X}[i])]$
  **else**
    $f_X[i] \leftarrow [\hat{X}[i],\ \hat{X}[i]^3, \dots, \hat{X}[i]^{2Z-3}, 0, 0, 0]$
  **end if**
**end for**
$Y \leftarrow f_X \cdot a^{\times}$
Reshape $Y$ back to $\mathbb{R}^{B\times C\times M\times N}$ **return** $Y$

element is true if the corresponding element of $\hat{X}$ exceeds the threshold $\delta$. Polynomial terms are calculated based on whether the condition $O$ is true or false: if true, the last two terms of $f_x$ are set to $\hat{X}$ and $-\delta \cdot sgn(\hat{X})$ respectively; if false, polynomial terms $x^{2k+1}$ for $k$ from 0 to $Z - 2$ are computed and set.

The final output tensor $Y$ is obtained by multiplying the matrix of polynomial terms $f_x$ with the vector of polynomial coefficients $a$, and reshaping the result back to the original size $B \times C \times M \times N$.

**Quaternion Hadamard Polynomial Denoising Block (QHPDB):** The QHPDB effectively suppresses adversarial noise by leveraging the Walsh-Hadamard Transform (WHT) and quaternion convolution. The process begins with applying the WHT to the input tensor and converting the data into the transform domain, where noise can be more easily identified and suppressed. For an input tensor $X \in \mathbb{R}^{B\times C\times M\times N}$, the 2D WHT is applied along the last two axes, resulting in $\hat{X} = \mathrm{WHT}(X)$. Then, quaternion convolution $QConv$ with learnable kernel $W_{st}$ is performed on $\hat{X}$ to replace the scaling operation. The transformed and scaled tensor $\hat{X}_{st} = QConv(\hat{X}, W_{st})$ undergoes polynomial thresholding to attenuate high-frequency components $\tilde{Y} = PT(\hat{X}_{st})$. After thresholding, the inverse WHT is applied to bring the data back to the spatial domain, yielding the tensor $Y = WHT^{-1}(\tilde{Y})$.

**Quaternion Denoising Residual Block (QDRB):** The block begins with a Quaternion Convolution layer that has a specific kernel size. Using quaternion convolutions is especially beneficial here because it effectively addresses the multidimensional characteristics of the data. After the initial convolution, the data flows through the QHPDB layer. Operating in the transform domain with the WHT, the QHPDB applies polynomial thresholding (PT). An additional branch carries the original features through a single quaternion convolution layer to ensure that key image features are preserved during denoising. This helps maintain important details that remain unaffected by noise removal. The block also sequentially integrates Channel Attention and Spatial Attention mechanisms ially.

**Channel Attention (CA):** selects the most informative feature channels by computing a channel-wise attention map and multiplying it with the input features. The CA mechanism is mathematically represented as follows:

$$CA(X) = \sigma\left(QConv2\left(ReLU\left(QConv1\left(AvgPool(X)\right)\right)\right)\right) \quad (12)$$

where $AvgPool(X)$ is the adaptive average pooling operation, reducing each channel to a single value, $QConv1$ is a quaternion convolution layer reducing the number of channels by the reduction ratio, ReLU is the $ReLU$ activation function, $QConv2$ is a quaternion convolution layer restoring the original number of channels, and σ is the sigmoid activation function producing the attention map.

**Spatial Attention (SA):** highlights significant spatial features by applying a series of convolutions and activations to enhance the regions of interest in the feature map. The SA mechanism is mathematically represented as follows:

$$\mathrm{SA(X)} = \sigma\left(\mathrm{QConv3}\left(\mathrm{ReLU}\left(\mathrm{QConv2}\left(\mathrm{QConv1(X)}\right)\right)\right)\right) \quad (13)$$

where $QConv1$ is the first quaternion convolution layer with a kernel size of 3x3, $QConv2$ is a second quaternion convolution layer reducing the number of channels, ReLU is the ReLU activation function, $QConv3$ is the final quaternion convolution layer restoring the original number of channels, and σ is the sigmoid activation function producing the attention map.

Finally, QDRB adds the input features back to the output. The whole process could be represented as follows:

$$\widehat{H}_1^n = \mathrm{QHPDB}\left(\mathrm{QConv1}(\mathrm{X}^{n-1}, W_1)\right) \quad (14)$$

$$\widehat{H}_2^n = \mathrm{QConv2}(X^{n-1}, W_2) \quad (15)$$

$$X^n = \mathrm{SA}\left(\mathrm{CA}\left(\widehat{H}_1^n + \widehat{H}_2^n\right)\right) + X^{n-1} \quad (16)$$

where $X^{n-1}$ is the input to the $n$-th QDRB, $\widehat{H}_1^n$ and $\widehat{H}_2^n$ are intermediate feature maps processed through the QHPDB and an additional QConv layer, respectively.

**Quaternion Feature Aggregation and Refinement Block (QFARB):** at the end of the processing, the feature map is adaptively refined following the procedure proposed in [39] and adapted for the quaternion case to robustly restore fine structural and textural details. The input features pass through a series of quaternion convolutional layers, efficiently capturing complex inter-channel relationships. The output undergoes global average pooling (GAP) to condense spatial information, followed by additional QConv layers and hyperbolic tangent (tanh) activations to refine the features. The attention map $A$ is generated using a sigmoid activation function on another quaternion convolution layer output. This map weights the original and refined features to select the most informative parts. The final output $\hat{Y}$ is computed as a weighted sum of these features, preserving essential details while enhancing image quality. The process within the QFARB is described by:

$$\hat{A}_1^n = tanh\left(\mathrm{QConv}\left(\mathrm{QConv}\left(\mathrm{GAP}(Y)\right)\right)\right) \quad (17)$$

$$\hat{A}_2^n = tanh(QConv(QConv((Y)))) \quad (18)$$

$$\hat{Y} = Y \odot \widehat{\mathrm{A}}_2 + \left(1 - \widehat{\mathrm{A}}_2\right) \odot \widehat{\mathrm{A}}_1 \quad (19)$$

where, $\hat{A}_1$ is the refined feature map, $\hat{A}_2$ is the attention map, and $\hat{Y}$ is the final output feature.

## IV. Dataset

To evaluate and train the QHNet, we have collected a custom dataset AWCVD covering diverse adverse weather conditions, including haze, rain, and snow. Our dataset was built by attacking images sampled from various synthetic datasets on different state-of-the-art models. For dehazing, we attacked DehazeFormer [40], MixDehazeNet [41], FSNet [42], DSANet [39], and Chen et al. [43] on RESIDE-6K [44] dataset. For rain-streak removal, we targeted M3SNet [45], Restormer [46], UDR-S2Former [47], and Chen et al. [43] on Rain-13k [48] dataset. For snow removal, we attacked DSANet [39], OKNet [49], and Chen et al. [43] on the CSD dataset [50]. These models were trained on the respective datasets and selected to represent a combination of CNN- and transformer-based approaches, ensuring a comprehensive evaluation.

We employed the Fast Gradient Sign Method (FGSM) and its iterative version (I-FGSM) as our first-order gradient methods to produce adversarial examples [51], [52]. An attack involves a loss function $\mathcal{L}(x_c + \rho, y_c; \theta)$, where θ denotes the network parameters. The aim is to maximize this loss by solving:

$$\rho = \arg \max_{\rho \in R^m} \mathcal{L}\,(x_c + \rho, y_c; \theta) \quad (20)$$

FGSM achieves this in a single step by determining adversarial perturbations. It does so by moving in the direction opposite to the gradient of the loss function with respect to the input (∇):

$$\mathrm{x_{adv}} = \mathrm{x_c} + \varepsilon \cdot \mathrm{sign}\left(\nabla\mathcal{L}(\mathrm{x_c}, \mathrm{y_c}; \theta)\right) \quad (21)$$

where, $\varepsilon$ represents the step size, which effectively bounds the $l_\infty$ norm of the perturbation.

I-FGSM applies the perturbation iteratively with the update rule:

$$\mathrm{x_{m+1}} = \mathrm{clip}\left(\mathrm{x_m} + \alpha \cdot \mathrm{sign}\left(\nabla\mathcal{L}(\mathrm{x_m}, \mathrm{y_c}; \theta)\right)\right) \quad (22)$$

where $m$ ranges from 0 to M, with $\mathrm{x_0} = \mathrm{x_c}$. After $M$ iterations, the final adversarial example is $\mathrm{x_{adv}} = \mathrm{x_M}$.

We use different combinations of $\varepsilon$ (2/255, 4/255, 6/255, 8/255, 10/255, and 15/255) and iteration counts ($i = 1, 3, 5, 7$, and 11) to attack the selected models. This approach enabled us to generate various adversarial examples paired with their clean counterparts for training our defense model. In total, we sampled 11,190 images for training, distributed as follows: 3000 from Rain-13k, 5000 from RESIDE-6K, and 3190 from CSD. For testing, we sampled 2100 images from the same "train" split of the original dataset, distributed as follows: 600 from Rain-13k, 1000 from RESIDE-6K, and 500 from CSD. All images were resized to match the size of the Test split of the dataset, which is used solely for validation during training and in ablation studies. The true effectiveness of the defense technique should be assessed using the testing datasets that come with the original datasets and the attack on the target model.

## V. Experiments

### *A. Experimental procedures*

We evaluate QHNet's performance in defending against adversarial attacks across three low-level computer vision tasks: haze removal, rain-streak removal, and snow removal.

We have attacked recent weather removal methods using FGSM ($\varepsilon = 2/255$), I-FGSM ($\varepsilon = 5/255$, $i = 5$), and I-FGSM ($\varepsilon = 5/255$, $i = 10$). Attacked images were processed by QHNet, by super-resolution technique ESRGAN [53], and by the state-of-the-art denoising method KBNet [54]. Then, we applied the target method to the original attacked image, and the images were processed with QHNet, ESRGAN, and KBNet.

TABLE II

SYNTHETIC HAZE REMOVAL RESULTS (RESIDE-6K DATASET)

| Attack Method | Dehazing method | Original/Attacked | | Super-resolution | | Denoising | | QHNet | |
|---|---|---|---|---|---|---|---|---|---|
| | | PSNR | SSIM | PSNR | SSIM | PSNR | SSIM | PSNR | SSIM |
| FGSM $\varepsilon$ =2/255 $i$=1 | DehazeFormer [40] | 26.208/ 20.954 | 0.954/ 0.898 | 21.094 | 0.893 | 21.051 | 0.904 | **22.830** | **0.921** |
| | MixDehazeNet [41] | 26.335/18.238 | 0.942/0.852 | 18.659 | 0.856 | 21.323 | 0.865 | **21.084** | **0.893** |
| | FSNet [42] | 27.231/19.607 | 0.947/0.873 | 19.844 | 0.874 | 22.476 | 0.872 | **22.076** | **0.900** |
| | DSANet [39] | 27.283/18.883 | 0.948/0.855 | 19.057 | 0.855 | 23.172 | 0.889 | **21.586** | **0.889** |
| | Chen et al. [43] | 29.284/22.557 | 0.970/0.915 | 23.176 | 0.920 | 25.366 | 0.924 | **26.031** | **0.952** |
| I-FGSM $\varepsilon$ =5/255 $i$=5 | DehazeFormer [40] | 26.208/9.187 | 0.954/0.628 | 9.863 | 0.649 | 19.236 | 0.837 | **22.076** | **0.919** |
| | MixDehazeNet [41] | 26.335/8.268 | 0.942/0.570 | 8.690 | 0.589 | 18.085 | 0.817 | **21.320** | **0.893** |
| | FSNet [42] | 27.231/10.233 | 0.947/0.432 | 11.018 | 0.481 | 22.388 | 0.872 | **23.941** | **0.913** |
| | DSANet [39] | 27.283/13.031 | 0.948/0.717 | 13.427 | 0.729 | 22.236 | 0.868 | **23.604** | **0.907** |
| | Chen et al. [43] | 29.284/12.695 | 0.970/0.697 | 13.107 | 0.711 | 21.288 | 0.872 | **25.532** | **0.947** |
| I-FGSM $\varepsilon$ =5/255 $i$=10 | DehazeFormer [40] | 26.208/7.728 | 0.954/0.570 | 8.343 | 0.592 | 19.873 | 0.842 | **23.692** | **0.932** |
| | MixDehazeNet [41] | 26.335/7.697 | 0.942/0.536 | 8.058 | 0.556 | 18.085 | 0.817 | **23.257** | **0.915** |
| | FSNet [42] | 27.231/6.752 | 0.947/0.167 | 7.348 | 0.206 | 22.414 | 0.869 | **24.873** | **0.922** |
| | DSANet [39] | 27.283/12.055 | 0.948/0.677 | 12.720 | 0.698 | 22.236 | 0.868 | **25.073** | **0.926** |
| | Chen et al. [43] | 29.284/10.962 | 0.970/0.623 | 11.586 | 0.651 | 20.835 | 0.854 | **27.237** | **0.957** |

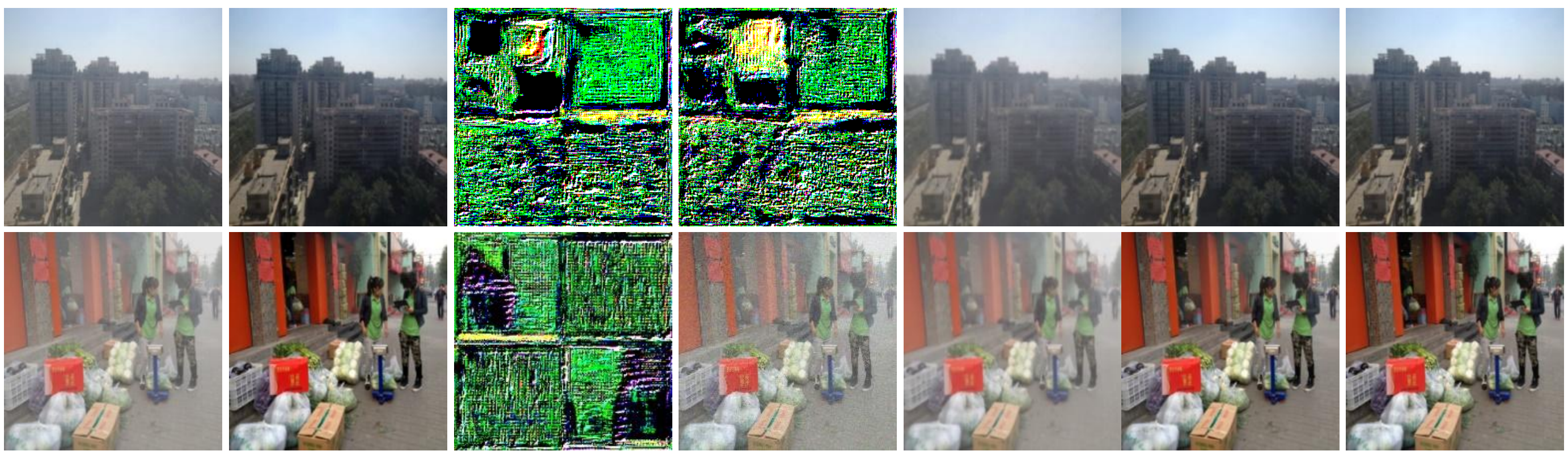

a) Input b) FSNet c) Attacked d) SR e) Denoising f) QHNet g) GT

**Fig. 5.** Haze removal by the FSNet method on the RESIDE-6K dataset. With I-FGSM attack, $\varepsilon = 5/255, i = 5$. (a) Input image; b) FSNet without attack - performs well (c) FSNet on attacked image - severe artifacts damage both images; (d) super-resolution can prevent artifacts in 1 out of 2 cases, but FSNet can still not remove the haze; (e) denoising prevents artifacts in all cases, but FSNet can still not remove the haze; (f) QHNet leads to the successful removal of haze on all images; (g) provides ground truth for comparison.

**Measuring defense efficiency:** We measured the quality of restoration using PSNR and SSIM, common metrics for checking image quality. The results are presented in Tables II-IV and Figures 5-7. For dehazing, we attacked DehazeFormer [40], MixDehazeNet [41], FSNet [42], DSANet [39], and Chen et al. [43] on RESIDE-6K [44] dataset. For rain-streak removal, we targeted M3SNet [45], Restormer [46], UDR-S2Former [47], and Chen et al. [43] on Rain-13k [48] dataset. For snow removal, we attacked DSANet [39], OKNet [49], and Chen et al. [43].

We also evaluate the classification setup by comparing it with recent purification-based methods, such as DiffPure [23], ADBM [24], and AdvPFY [35].

TABLE III

HEAVY RAIN REMOVAL RESULTS (RAIN100H DATASET)

| | | Original/Attacked | | Super-resolution | | Denoising | | QHNet | |
|---|---|---|---|---|---|---|---|---|---|
| Attack Method | Rain-removal method | PSNR | SSIM | PSNR | SSIM | PSNR | SSIM | PSNR | SSIM |
| FGSM $\varepsilon$ =2/255 $i$=1 | M3SNet [45] | 29.307/28.135 | 0.928/0.920 | 27.552 | 0.909 | 18.916 | 0.716 | **28.572** | **0.923** |
| | Restormer [46] | 29.584/28.024 | 0.932/0.923 | 27.824 | 0.914 | 18.741 | 0.694 | **28.861** | **0.928** |
| | UDR-S2Former [47] | 19.486/19.505 | 0.753/0.750 | 19.315 | 0.748 | 15.887 | 0.611 | **19.606** | **0.752** |
| | Chen et al. [43] | 25.929/25.216 | 0.886/0.876 | 25.064 | 0.872 | 18.147 | 0.673 | **25.582** | **0.882** |
| I-FGSM $\varepsilon$ =5/255 $i$=5 | M3SNet [45] | 29.307/18.195 | 0.928/0.801 | 20.111 | 0.836 | 18.972 | 0.715 | **25.253** | **0.905** |
| | Restormer [46] | 29.584/18.797 | 0.932/0.805 | 20.734 | 0.844 | 18.691 | 0.722 | **26.504** | **0.916** |
| | UDR-S2Former [47] | 19.486/17.506 | 0.753/0.673 | 17.743 | 0.683 | 15.974 | 0.606 | **18.844** | **0.714** |
| | Chen et al. [43] | 25.929/18.447 | 0.886/0.764 | 19.433 | 0.791 | 17.809 | 0.690 | **23.622** | **0.866** |
| I-FGSM $\varepsilon$ =5/255 $i$=10 | M3SNet [45] | 29.307/14.547 | 0.928/0.690 | 16.833 | 0.764 | 18.643 | 0.700 | **26.211** | **0.909** |
| | Restormer [46] | 29.584/15.194 | 0.932/0.703 | 18.073 | 0.790 | 18.691 | 0.722 | **27.220** | **0.919** |
| | UDR-S2Former [47] | 19.486/15.610 | 0.753/0.605 | 16.061 | 0.624 | 15.698 | 0.592 | **18.761** | **0.713** |
| | Chen et al. [43] | 25.929/15.036 | 0.886/0.635 | 16.447 | 0.698 | 17.517 | 0.680 | **24.020** | **0.869** |

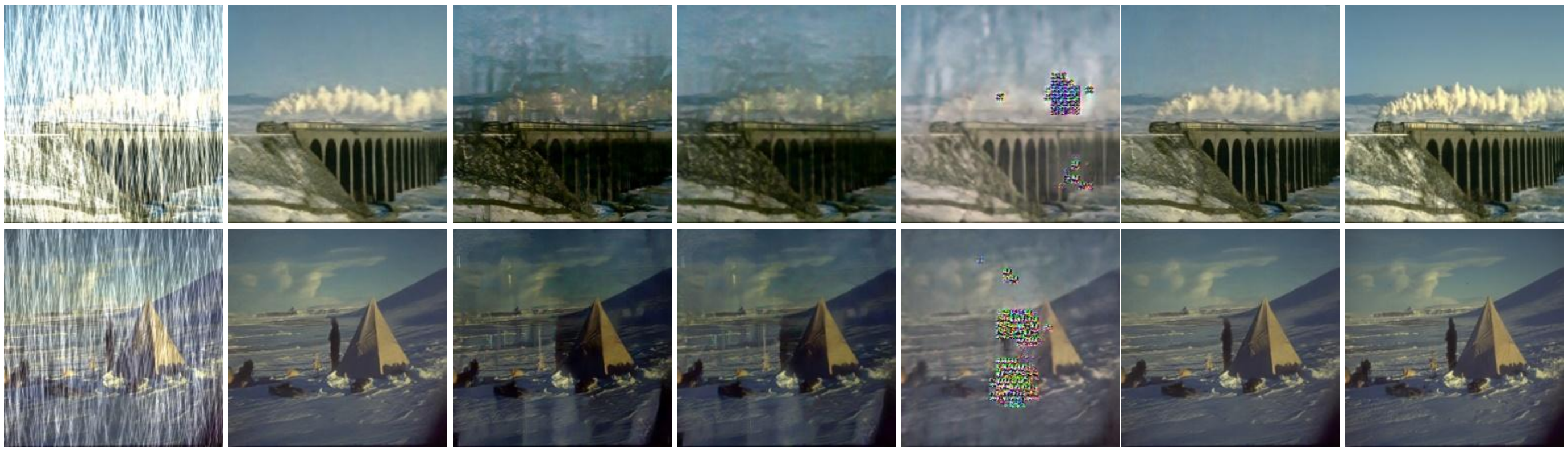

a) Input b) M3SNet c) Attacked d) SR e) Denoising f) QHNet g) GT

**Fig. 6.** Rain-streak removal by M3SNet on Rain100H dataset. With the I-FGSM attack, $\epsilon = 5/255, i = 10$. (a) Input image; (b) M3SNet non-attacked input image; (c) M3SNet on attacked image: failing to remove streaks, with added artifacts; (d) Super-resolution; (e) denoising does not improve the situation significantly; (f) QHNet reduces effects of attack; (g) Ground truth.

To evaluate the effect of gradient masking, we use the same PGD–BPDA–EOT protocol and set budgets across all three baselines, utilizing their publicly available inference pipelines and recommended checkpoints. [33], [52], [55] Each method's default forward computation remains unchanged (e.g., deterministic sampling where available), and BPDA is used only during the backward pass to compute gradients with respect to the original input. It also acts as the identity function for non-differentiable steps in the baselines and employs QHNet's polynomial surrogate in our module. This approach ensures a fair, defense-aware comparison across purification methods.

### *B. Implementation details*

The model is trained on 64x64 image patches, leveraging the AdamW optimizer with parameters $\beta_1 = 0.9, \beta_2 = 0.999$. The learning rate is set to an initial value of $1 \times 10^{-3}$, decaying to a minimum of $1 \times 10^{-7}$ through a cosine annealing schedule with a warm-up phase of 2 epochs. This training strategy ensures a smooth and effective learning process. The training process spans 250 epochs with a batch size of 12, conducted on a single NVIDIA A100 GPU. We use the Structural Similarity Index (SSIM) loss function:

$$\mathcal{L} = 1 - SSIM(QHNet(X), Y) \tag{23}$$

where $QHNet(\cdot)$ is the proposed network, $X$ represents the attacked input image, and $Y$ is the ground truth image.

### *C. Experimental results*

In this subsection, we discuss the effectiveness of QHNet in protection against adversarial attacks.

**Haze removal:** Table II and Fig. 5 present results for attacking haze removal methods: DehazeFormer, MixDehazeNet, FSNet, DSANet, and Chen et al. For haze removal techniques, even FGSM with $\varepsilon$=2/255 significantly reduces performance (PSNR from 26.208 to 20.954, and SSIM from 0.954 to 0.898 for DehazeFormer). Super-resolution introduces artifacts and generally offers only a slight improvement. Denoising performs better, but QHNet significantly improves the target model's performance on attacked images. The performance of all dehazing methods is severely affected by the I-FGSM attack with $\varepsilon$=2/255 and $i = 10$. For example, for Chen et al., PSNR degrades from 29.284 to 10.962 and SSIM from 0.970 to 0.623. QHNet restores PSNR to 27.237 and SSIM to 0.957, which is lower than the performance without an attack but still reasonable for subsequent computer vision applications, and significantly better than denoising and super-resolution improvements.

**Rain-streak removal:** Table III and Fig. 6 demonstrate the attack on rain-streak removal methods (M3SNet, Restormer, UDR-S2Former, Chen et al.) for the Rain100H dataset.

Light attacks ($\varepsilon$=2/255) do not significantly impact performance, but severe attacks ($\varepsilon$=5/255, i=10) drastically reduces performance. From ~30 PSNR to ~15 PSNR, both super-resolution and denoising fail to prevent degradation of rain-streak removal performance and introduction of artifacts. QHNet significantly reduces degradation, especially with DSANet.

TABLE IV: SNOW REMOVAL RESULTS (CSD DATASET)

| Attack method | Snow-removal method | Original results and attack | | Super-resolution | | Denoising | | QHNet | |
|---|---|---|---|---|---|---|---|---|---|
| | | PSNR | SSIM | PSNR | SSIM | PSNR | SSIM | PSNR | SSIM |
| FGSM $\varepsilon$ =2/255 $i$=1 | DSANet [39] | 29.038/13.304 | 0.941/0.669 | 14.519 | 0.697 | 22.343 | 0.859 | **28.491** | **0.935** |
| | OKNet [49] | 29.084/12.359 | 0.942/0.407 | 17.250 | 0.741 | 22.979 | 0.843 | **24.626** | **0.828** |
| | Chen et al. [43] | 26.749/21.277 | 0.920/0.868 | 22.008 | 0.879 | 23.702 | 0.883 | **23.826** | **0.899** |
| I-FGSM $\varepsilon$ =5/255 $i$=5 | DSANet [39] | 29.038/8.659 | 0.941/0.222 | 12.013 | 0.452 | 18.579 | 0.799 | **28.470** | **0.936** |
| | OKNet [49] | 29.084/5.470 | 0.942/0.015 | 5.477 | 0.015 | 21.457 | 0.833 | **24.624** | **0.829** |
| | Chen et al. [43] | 26.749/14.042 | 0.920/0.717 | 14.208 | 0.728 | 18.697 | 0.824 | **24.303** | **0.901** |
| I-FGSM $\varepsilon$ =5/255 $i$=10 | DSANet [39] | 29.038/7.369 | 0.941/0.142 | 11.184 | 0.387 | 17.615 | 0.781 | **28.527** | **0.936** |
| | OKNet [49] | 29.084/5.469 | 0.942/0.015 | 5.472 | 0.015 | 21.002 | 0.814 | **24.638** | **0.829** |
| | Chen et al. [43] | 26.749/13.049 | 0.920/0.677 | 13.344 | 0.693 | 17.552 | 0.801 | **25.679** | **0.911** |

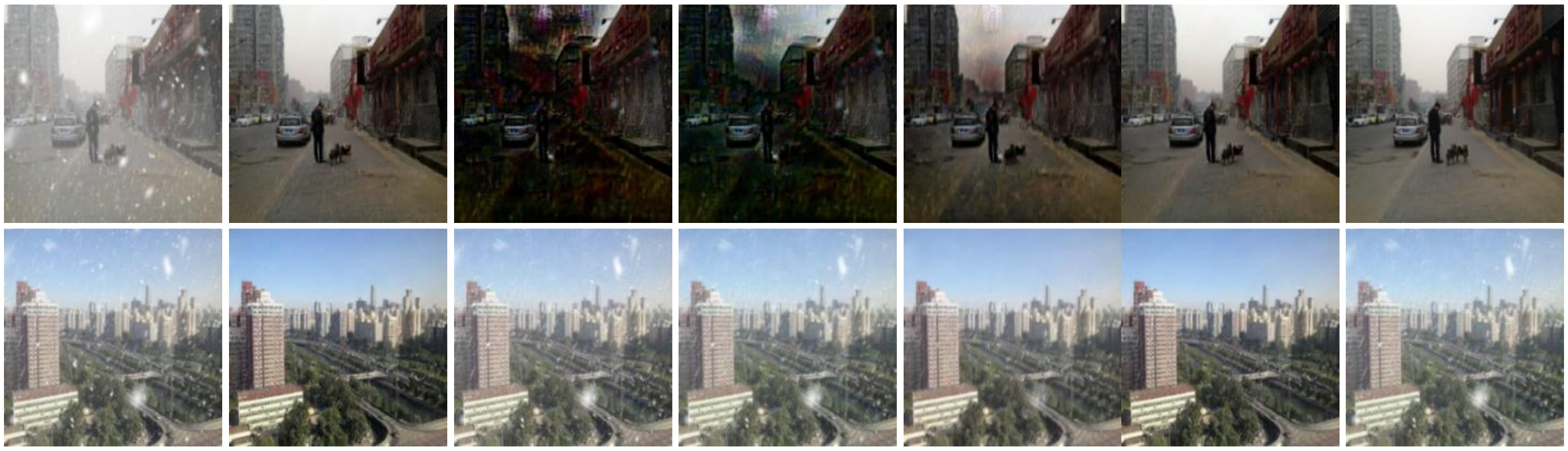

**Fig. 7.** Snow removal by Chen et al. on the CSD dataset. With the I-FGSM attack, $\varepsilon = 5/255, i = 5$. (a) Input image; (b) Non-attacked image restored by Chen et al.; (c) Attacked image restored by Chen et al. with severe artifacts or unremoved snowflakes; (d) Super-resolution and denoising; (e) Improve quality but introduce artifacts and darken the image; (f) QHNet successfully removes snowflakes, producing images close to the ground truth; (g) Ground truth.

Light attacks ($\varepsilon$=2/255) do not significantly impact performance, but severe attacks ($\varepsilon$=5/255, i=10) drastically reduces performance. From ~30 PSNR to ~15 PSNR, both super-resolution and denoising fail to prevent degradation of rain-streak removal performance and introduction of artifacts. QHNet significantly reduces degradation, especially with DSANet.

**Snow removal:** Table IV and Fig. 7 present snow removal results. Methods like DSANet and OKNet work well under normal conditions, but degrade significantly under FGSM ($\varepsilon$=2/255) attacks. Super-resolution and denoising methods do not fully fix the damage and often add artifacts. QHNet achieves the highest PSNR and SSIM scores, effectively recovering attacked images. Overall, super-resolution and denoising methods do not fully repair damage and often introduce artifacts. QHNet consistently achieves the highest PSNR and SSIM scores, effectively recovering images from attacks.

*D. Adaptive White-Box Evaluation (PGD–BPDA–EOT)*

We perform a comprehensive, defense-aware evaluation within a white-box threat model, where the attacker has full knowledge of the classifier, purification module, and all processing components. Because gradient-based attacks pose the greatest white-box threat, they are adapted to account for all parts. The attacker employs direct gradients to craft highly effective adversarial samples. Gradients are backpropagated through non-differentiable components using Backward Pass Differentiable Approximation (BPDA), while stochasticity is handled with Expectation over Transformation (EOT). To maintain fairness, all defenses are tested with deterministic forward passes, with EOT used solely by the attacker. For baseline methods, identity-BPDA is applied through non-differentiable steps. Our approach, however, replaces these steps with a smooth polynomial surrogate during backward passes, leaving the forward process unchanged.

**Evaluation Protocol and Baselines:** Evaluation employs $\ell_\infty$-Projected Gradient Descent (PGD) attacks on clean-trained ResNet-18 classifiers. The attack configuration uses 20 iterations, a step size of 2/255, and EOT with 20 Monte-Carlo samples per gradient estimate under consistent perturbation budgets. This substantial adaptive setting requires 400 forward evaluations per attack, aligning with recent purifier assessment recommendations. We compare our method against three representative purification approaches: (i) DiffPure [23]: Performs forward diffusion followed by reverse SDE denoising, with gradient backpropagation through the reverse process; (ii) ADBM [24]: Employs learned diffusion bridges that map diffused adversarial inputs toward clean manifolds, and (iii) AdvPFY [35]: Utilizes variational autoencoder-style manifold projection with semantic consistency objectives designed for defense-aware attack resilience.

TABLE V
CIFAR-10 ε-SWEEP UNDER PGD-BPDA-EOT
TOP-1 ACCURACY (%). HIGHER IS BETTER.

| Method | Clean | 1/255 | 2/255 | 4/255 | 8/255 |
|---|---|---|---|---|---|
| ResNet-18 (no defense) | 95.2 | 68.3 | 42.1 | 18.2 | 3.8 |
| DiffPure [23] | 93.8 | 74.6 | 56.2 | 28.9 | 8.4 |
| ADBM [24] | 93.5 | 75.8 | 58.3 | 31.2 | 9.1 |
| AdvPFY [35] | 93.9 | 76.2 | 59.1 | 32.4 | 9.6 |
| QHNet → ResNet-18 | **94.3** | **78.4** | **62.7** | **36.1** | **11.3** |

TABLE VI
CIFAR-100 ε-SWEEP
PGD-BPDA-EOT (k=20).

| Method | Clean | 1/255 | 2/255 | 4/255 | 8/255 |
|---|---|---|---|---|---|
| ResNet-18 (no defense) | 77.1 | 38.7 | 19.3 | 5.2 | 0.8 |
| DiffPure [23] | 75.9 | 44.2 | 26.8 | 9.7 | 1.6 |
| ADBM [24] | 75.7 | 46.1 | 28.4 | 10.9 | 2.0 |
| AdvPFY [35] | 76.1 | 46.8 | 29.2 | 11.3 | 2.2 |
| QHNet → ResNet-18 | **76.4** | **49.3** | **31.6** | **12.8** | **2.7** |

**CIFAR-10/100 Evaluation:** On CIFAR-10, all methods show the expected monotonic accuracy drop as perturbation budgets increase (Table V), confirming proper adaptive attack tuning. While undefended models fail quickly, purification methods significantly improve robustness. QHNet is consistently strongest, reaching 11.3% robust accuracy at $\ell_\infty = 8/255$, a 1.7-point gain over AdvPFY, with similar clean accuracy. The advantage persists on CIFAR-100 (Table VI), where QHNet attains 2.7% robust accuracy versus 2.2% for AdvPFY.

**Gradient-Obfuscation Check:** To verify the absence of masking, we examine attack ordering at $\epsilon = 8/255$. Table VII confirms the canonical white-box ≤ transfer ≤ black-box pattern. The score-based Square Attack [56] is applied end-to-end under a fixed query budget, with all purifiers evaluated using a single deterministic forward pass (e.g., DDIM for DiffPure, a single VAE pass for AdvPFY), thereby exposing no gradients or internal states.

**Computational Cost:** Table VIII summarizes parameter counts and median end-to-end runtime per 64×64 RGB sample across ≥100 runs. QHNet offers strong robustness with substantially lower computational cost than diffusion-based purifiers..

TABLE VII: ACCURACY (%) AT $\varepsilon = 8/255$ ON CIFAR-10 ACROSS ATTACK FAMILIES

| Method | White-box (PGD-BPDA-EOT) | Transfer | Black-box (SQUARE) |
|---|---|---|---|
| ResNet-18 (no defense) | 0.4 | 2.8 | 10.6 |
| DiffPure [23] | 3.2 | 8.9 | 18.4 |
| ADBM [24] | 3.8 | 9.6 | 19.2 |
| AdvPFY [35] | 4.1 | 10.2 | 19.8 |
| QHNet → ResNet-18 | **4.9** | **10.8** | **20.6** |

Diffusion-style baselines are shown with their reverse-process step counts to demonstrate sampling effort. This summary clarifies that, while diffusion-based approaches are highly robust, they require many reverse steps and thus take notably more wall-clock time per sample. In contrast, our module operates in a single pass and achieves accuracy comparable to these more computationally intensive methods. When considered alongside the computational footprint implied by EOT-averaged PGD, these results suggest that our approach provides meaningful, diffusion-level robustness while maintaining the simplicity and efficiency of a single-pass purifier.

TABLE VIII: SIZE AND COMPUTE COST FOR PURIFIER AT 64X64

| Method | # parameters (M) | Reverse steps | Time per sample (ms) |
|---|---|---|---|
| ResNet-18 (no defense) | 11.2 | - | 0.8 |
| DiffPure (DDPM) [23] | 52.6 | 100 | 485 |
| DiffPure (DDIM) [23] | 52.6 | 10 | 52 |
| ADBM [24] | 48.3 | 50 | 245 |
| AdvPFY [35] | 18.7 | - | 8.2 |
| QHNet → ResNet-18 | 14.9 | - | 3.1 |

### *E. Ablation study*

In this section, we analyze the effectiveness of the components of the proposed architecture, QHNet. The "Real" network, which serves as the baseline, has the same architecture as QHNet but does not utilize the quaternion approach. It contains 16.8 million parameters and achieves lower PSNR (43.1448) and SSIM (0.9894) than QHNet.

TABLE IX: ABLATION STUDY ON THE COMPONENTS OF THE PROPOSED ARCHITECTURE QHNET

| QHPDB | QFARB | Attention | PT | #params | PSNR | SSIM |
|---|---|---|---|---|---|---|
| × | ✓ | ✓ | ✓ | 4,576,392 | 43.2330 | 0.9907 |
| ✓ | × | ✓ | ✓ | 3,649,296 | 42.8338 | 0.9899 |
| ✓ | ✓ | × | ✓ | 2,648,498 | 42.8358 | 0.9890 |
| ✓ | ✓ | ✓ | × | 3,676,104 | 43.2343 | 0.9907 |
| ✓ | ✓ | ✓ | ✓ | 3,676,104 | 43.3087 | 0.9934 |

As shown in Table IX, removing any component from QHNet results in a performance drop, confirming the contribution of each module. QHNet, which includes all components (QHPDB, QFARB, Attention, and PT), achieves the best results with a PSNR of 43.3087 and an SSIM of 0.9934, demonstrating the effectiveness of integrating all these developed modules.

## VI. CONCLUSION

This paper presents QHNet, a lightweight, model-independent purification defense against first-order white-box attacks for adverse-weather image restoration. While recent deraining, desnowing, and dehazing networks remain highly susceptible to gradient-aligned perturbations, QHNet offers an efficient alternative to computationally expensive defenses such as adversarial training and diffusion-based purifiers. QHNet combines quaternion processing with polynomial thresholding through the QHPDB and QDRB modules, enabling effective suppression of adversarial noise within an encoder–decoder framework. Experiments across haze, rain, and snow datasets demonstrate consistent robustness improvements, validated by PSNR and SSIM gains, and highlight the detrimental impact of attacks on both restoration quality and downstream perception. Future work will extend QHNet to support black-box and gray-box threat models and explore quaternion-based probabilistic reasoning (QPN) to further enhance resilience in complex, uncertain real-world conditions.

## ACKNOWLEDGEMENT

This research was supported by the Google Fellowship Grant, project 7E263-05-02. The authors gratefully acknowledge this financial support, which enabled the work.